# Hand tracking for clinical applications: validation of the Google MediaPipe Hand (GMH) and the depth-enhanced GMH-D frameworks.

Gianluca Amprimo[a,b,**], Giulia Masi[c], Giuseppe Pettiti[b], Gabriella Olmo[a], Lorenzo Priano[c,d] and Claudia Ferraris[b,*]

[a]Politecnico di Torino - Control and Computer Engineering Department, Corso Duca degli Abruzzi, 24, Turin, 10129, Italy
[b]National Research Council - Institute of Electronics, Information Engineering and Telecommunications, Corso Duca degli Abruzzi, 24, Turin, 10029, Italy
[c]Università di Torino - Neurosciences Department "Rita Levi Montalcini", Via Verdi, 8, Turin, 10124, Italy
[d]Istituto Auxologico Italiano – Department of Neurology and Neurorehabilitation S. Giuseppe Hospital, Via Cadorna 90, Oggebbio, 28824, Italy

## ARTICLE INFO

## ABSTRACT

Accurate 3D tracking of hand and fingers movements poses significant challenges in computer vision. The potential applications span across multiple domains, including human-computer interaction, virtual reality, industry, and medicine. While gesture recognition has achieved remarkable accuracy, quantifying fine movements remains a hurdle, particularly in clinical applications where the assessment of hand dysfunctions and rehabilitation training outcomes necessitate precise measurements. Several novel and lightweight frameworks based on Deep Learning have emerged to address this issue; however, their performance in accurately and reliably measuring fingers movements requires validation against well-established gold standard systems. In this paper, the aim is to validate the hand-tracking framework implemented by Google MediaPipe Hand (GMH) and an innovative enhanced version, GMH-D, that exploits the depth estimation of an RGB-Depth camera to achieve more accurate tracking of 3D movements. Three dynamic exercises commonly used by clinicians to assess hand dysfunctions, namely Hand Opening-Closing, Single Finger Tapping and Multiple Finger Tapping are considered. Results demonstrate high temporal and spectral concordance of both frameworks with the gold standard. However, the enhanced GMH-D framework exhibits superior accuracy in spatial measurements compared to the baseline GMH, for both slow and fast movements. Overall, our study contributes to the advancement of hand tracking technology, the establishment of a validation procedure as a good-practice to prove efficacy of deep-learning-based hand-tracking, and proves the effectiveness of GMH-D as a reliable framework for assessing 3D hand movements in clinical applications.

## 1. Introduction

Monitoring, recognising, and interpreting the natural movement of the body, without the aid of devices and instrumentation that can alter its characteristics, are among the most currently addressed research topics in Computer Vision (CV) [59, 101]. In fact, it opens a multitude of possible scientific and consumer applications [96, 81]. Human Pose Estimation (HPE) and Human Action Recognition (HAR) are finding their way into the fields of human-computer interaction, virtual reality, robotics, sports, video surveillance, industry, biomechanics, and medicine [74, 6, 22, 23, 7, 59, 103, 62, 16]. However, despite advances in the accurate recognition and estimation of static or quasi-static poses and gestures [76], there is still a long way to go in tracking and measuring motion characteristics, mainly when focusing on small body parts such as the hand and fingers [4, 12].

The development of real-time, robust, non-invasive, cost-effective, and accurate algorithms for tracking human hand and finger movements has attracted much interest in recent years. However, this is still an open challenge [66]. Meeting all the mentioned requirements is complex, and constraints are often established according to specific needs [78, 25]. Indeed, the human hand has a complex and fully articulated anatomical structure that allows for a myriad of coarse actions and fine movements, making its complete reconstruction and reliable analysis tough [24].

After several attempts to develop solutions for constrained hand tracking using devices and supporting aids (such as instrumented gloves), the first bare-hand solutions were proposed [83, 69, 95, 77, 87]. In most cases, these solutions were limited to offline processing or low frame rate, thus preventing their practical use where real-time motion capture and analysis are required, such as for clinical applications.

Recent and innovative methodologies, such as Deep Learning (DL) and pre-trained models, exhibit a great potential in the implementation of open-source and lightweight frameworks for motion capture. However, regarding spatial accuracy in measuring three-dimensional (3D) movements and their features, these solutions still have limitations in

*Corresponding author
**Principal corresponding author
📧 gianluca.amprimo@polito.it (G. Amprimo)
🌐 https://www.polito.it/personale?p=gianluca.amprimo (G. Amprimo); https://www.researchgate.net/profile/Giulia-Masi-2 (G. Masi); https://www.ieiit.cnr.it/people/Pettiti-Giuseppe (G. Pettiti); https://www.sysbio.polito.it/analytics-technologies-health/ (G. Olmo); https://neuroen.campusnet.unito.it/do/docenti.pl/Alias?lorenzo.priano#tab-profilo (L. Priano); https://www.ieiit.cnr.it/people/Ferraris-Claudia (C. Ferraris)

ORCID(s): 0000-0003-4061-8211 (G. Amprimo); 0000-0003-1940-2934 (G. Masi); 0000-0003-0547-0143 (G. Pettiti); 0000-0002-3670-9412 (G. Olmo); 0000-0002-7012-6501 (L. Priano); 0000-0001-5381-4794 (C. Ferraris)

clinical applications, where the assessment of motor performance requires high accuracy. In order to achieve objective evidence of their precision, it is necessary to compare the performance of such frameworks against traditional reference systems for human movement analysis, such as motion capture systems, rather than other manual instrumentation [46].

This paper presents a validation procedure against a gold standard system of both Google MediaPipe Hand (GMH) basic version [105] and GMH-D [9]. This latter is an innovative and enhanced framework for hand tracking based on GMH and an RGB-Depth (RGB-D) camera, that enables simultaneous, calibrated, and aligned colour and depth streams. Microsoft Azure Kinect (MAK) camera [1] was used in this study; however, GMH-D is also suitable for other optical sensors able to provide both colour and depth streams simultaneously. Three challenging tasks, commonly used in clinical examinations and rehabilitation of hand motor functions (e.g., for Parkinsonian subjects), were considered to compare 3D trajectories and estimated kinematic parameters, and validate the performance of the two frameworks. The main contributions of this work are the following:

- to validate the performance of basic GMH and enhanced GMH-D frameworks against a motion capture system, in terms of 3D trajectories and estimated spatial, temporal, and spectral features;

- to compare the performance of both frameworks in tracking hand movements during the three selected dynamic exercises;

- to establish good-practice guidelines for the validation of marker-less, deep-learning-based frameworks for hand tracking for clinical applications, since most of the current solutions lack a proper validation as measurement system (see Section 2.3).

The paper is organised as follows. Section 2 provides an overview of recent technologies and frameworks related to hand tracking, with focus on solutions for clinical scenarios. Section 3 describes the validation protocol and experimental setup. Section 4 presents the methodological approach to analyse and validate the frameworks. Section 5 reports and discusses the main results of the experimental tests. Finally, in Section 6 some concluding remarks are provided.

## 2. Background

In this Section, an overview of CV and DL approaches for hand tracking using RGB and RGB-D cameras is reported, focusing on their clinical applications. Then, GMH and its depth-enhanced version (GMH-D) are presented, before delving into the investigation of their validity for the clinical scenario of hand dexterity assessment.

### 2.1. Computer vision and hand tracking

Before the advent of DL, hand tracking for motion analysis heavily relied on traditional CV approaches, based on specialised image processing techniques and devices. Although many works employ wearable devices or sensorised gloves/exoskeletons [56, 71, 63, 88], non-contact motion tracking through vision systems has attracted great interest as it may overcome the main limitations of contact-based solutions: interfering effects on natural movements, discomfort due to wires, and bulkiness.

Traditional CV techniques have been extensively investigated to capture bare-hand movements from videos, including skin colour segmentation and mean shift algorithms with its variants [72, 99]. The availability of early RGB-D sensors quickly led to a more comprehensive 3D analysis by exploiting the potential of distance estimation through depth maps [104, 18, 21].

In other studies, passive gloves have supported the tracking of specific areas of the hand and fingers, solving skin colour issues, and reducing tracking complexity [61, 44, 39, 94, 27]. More recently, some optical devices (Leap Motion, Intel Real-Sense) have made available the first full 3D hand skeletal models, which have been successfully applied in various clinical studies [13, 41, 85, 11], albeit with specific performance-related constraints [28, 32].

### 2.2. Deep Learning for in-the-wild hand tracking

DL approaches for hand tracking from *in-the-wild* video sequences can be organised in a taxonomy, according to their input modality [17]: RGB, depth map, or mixed RGB-D.

Depth approaches were introduced to allow 3D reconstruction of the hand, following the increase in market-availability of depth sensors. Depth reconstruction provides several benefits: good shape information, insensitivity to shadow and illumination, and robustness to clutter. Convolutional Neural Networks (CNN) process this data to extract hand tracking information [90, 70, 58], possibly enforcing kinematics-based rules to improve the estimation [106]. Even if accurate, depth-methods have downsides such as large energy consumption, poor form factor, poor near-distance coverage, and difficult outdoor usage due to light interaction with Time of Flight (TOF) technology [49].

Considering multimodal methods (RGB-D), either the RGB stream is used for identifying the hand in two dimensions and then the associated depth stream is used to uplift joints [66], or the two modalities are fused to perform a single-shot estimation [40, 79]. Even when not employed for inference, the mixed RGB-D modality is frequently included at least in the training of most recent models working on RGB inference alone [17], to improve the quality of the final prediction.

Concerning RGB input modality, bi-dimensional (2D) hand landmarks extraction is usually embedded in many state-of-the-art HPE estimation methods such as OpenPose [15] and AlphaPose [26]. Few approaches, instead, concentrate on tracking exclusively the 2D hand and propose dedicated architectures [80, 30]. Indeed, the 2D representation alone does not allow to estimate complex movements that cannot be approximated on a planar projection. This becomes relevant in the case of fingers motion, due the numerous degrees of freedom to consider. A possible solution for inferring the third dimension after 2D pose extraction requires multiple-camera setups to perform geometrical triangulation [53]. However, this solution increases the complexity and the cost of the final acquisition system.

Several recent works have focused on the challenging task of directly estimating 3D coordinates from depth clues in monocular RGB videos. Since the first work from Zimmerman et al. [107], many architectures have been investigated [65, 35, 34]. However, these works report little information about the efficiency [86, 82], or claim real-time performance (>30 fps) without providing code to reproduce the results [14]. Moreover, top-tier accuracy on benchmark datasets is generally achieved by exploiting high-performance GPUs, which makes these solutions infeasible for applications outside research laboratories.

## 2.3. DL hand tracking in clinical applications

Despite the numerous works investigating hand tracking from a single RGB, depth or RGB-D camera, only a few solutions implement novel, *ad hoc* models [102, 97, 100, 92, 5, 54], whereas most clinical studies utilise well-established tools such as OpenPose [64, 52, 73, 48, 57, 29, 47, 75, 37], GMH [33, 51, 10, 31], and DeepLabCut [98, 68, 50, 38, 84, 93]. This latter is an easy-to-use framework for data labelling in videos and virtual-joints-regression model training, exploiting deep architectures and transfer learning [60].

Works exploiting OpenPose and DeepLabCut use 2D joints estimation, providing an evaluation of motions limited to specific parameters (e.g., frequency, angles between joints) that are easily retrievable from a planar view. In the case of custom DL models, they often employ pre-trained backbone networks to infer 3D coordinates on general-purpose hand tracking datasets and are fine-tuned on data collected for the specific clinical study. These models are solely validated based on their ability to provide predictions of impairment status consistent with the clinical examination [102, 97, 100, 92, 5, 54]. However, this does not provide validation of the DL model as an actual measurement tool. The results may be biased by the data collected specifically for the study, and the fitting of the classification/regression model, rather than the tracking quality in providing objectively correct measurements. Furthermore, these methods typically require GPU acceleration, often with low throughput (<30 fps) during joints estimation.

The evident gap between theoretical hand tracking using DL and its practical application to the medical field seems to be caused by the complexity of the best performing models, which can be hardly applied in practice. Moreover, it must be considered that clinical applications often do not require real *in-the-wild* capability of tracking (i.e., videos are often standardised, within a controlled scenarios). Considering this factor, complexity of the DL model could be reduced, while focusing on improving its accuracy.

Additionally, the methods used in the clinical domain lack a proper validation procedure. Indeed, the employed methods miss measurement capabilities description in the specific application. For instance, in the case of OpenPose-based methods, it must be considered that the model was not specifically trained to recognise clinical assessment tasks (e.g., finger tapping, prono-supination, opening and closing of the hand), and this could impair accuracy when processing these specific activities. Only one study [29] validated Open-Pose performance in finger tapping task by analysing joint angles during task execution. Another study [50] validated 2D-poses given by DeepLabCut with respect to movement frequency using inertial sensors. A similar approach, focusing on resting tremor in Parkinson's disease, was done by [33] for GMH. However, no 3D tracking-based work provides performance validation compared to video-based motion capture, which is considered the gold standard for human motion analysis. This type of validation is especially significant for 3D approaches, since it allows to perform a broader comparison including relevant parameters such as spatial excursions, temporal duration and variation of the single movements, in terms of overall frequency of motion.

Moreover, the validation of the DL framework as a measurement tool is crucial to prove that subsequent stages, i.e- automatic disease/impairment assessment, are actually identifying a clinically-relevant evidence, based on objective alterations in the subjects' parameters. This is also essential to increase acceptability by clinical personnel, who is often hesitant to use approaches for which a high degree of interpretability/trust cannot be provided.

## 2.4. A focus on GMH and GMH-D

An interesting DL approach based on RGB input is included in MediaPipe [105], the solution for lightweight and portable Machine Learning (ML) pipelines by Google LLC. The GMH framework is composed of two sub modules: a Palm Detection (PD) and a Hand Landmarks Detection (HLD) models. First, the region of interest corresponding to the hand is identified using the PD model. Then, the HLD model bundle detects the 21 keypoints corresponding to real joints within the detected hand regions. Figure 1 summarises the coordinate systems provided by GMH. As it can be observed, the framework provides both *Image Coordinates* in pixels, coupled with a dimensionless parameter $z_{j,im}$ that estimates the relative depth of joint $j$ with respect to the wrist reference, and a set of 3D *World Coordinates*, expressed in metres and centred in the bounding box of the palm detected by PD.

GMH was trained on approximately 30K real-world images, as well as several rendered synthetic hand shapes superimposed on various backgrounds to increase robustness. Moreover, it supports a frame rate in excess of 50 fps on a Google Pixel 6 phone using CPU only, or even faster (>80 fps) exploiting GPU acceleration, as reported by the pipeline official webpage [3].

This framework represents an interesting approach to hand tracking, balancing accuracy and time efficiency. This aspect is indeed crucial to develop easy-to-use and widely employable assessment systems. Regarding its application to the medical field, attempts have already been made to apply GMH to identify [96] and measure [33] resting tremor in Parkinson's disease, employing accelerometers to validate its performance. Moreover, the usage of GMH to measure fingers excursion from static images, in terms of relative joint-angles, was also proposed in [31] and validated using standard manual goniometry. However, it must be pointed out that such validation was carried out on single images, removing the complexity deriving from dynamic motion estimation and employing a protocol with several restrictions on hand positioning and environmental conditions. This limits the validity of the results to a very narrow scenario.

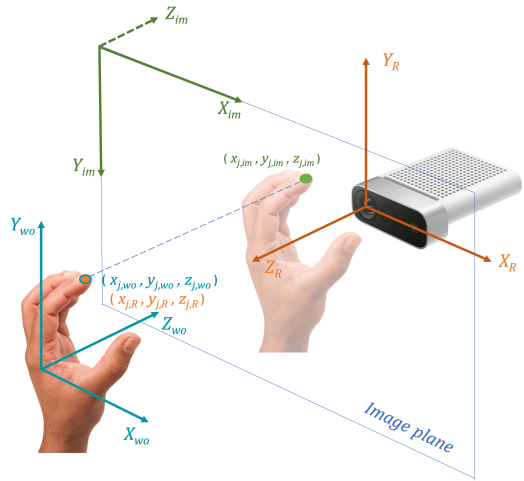

**Figure 1:** Set of coordinates tracked by GMH and GMH-D: for GMH, in green *Image Coordinates* (pixels), centred in the upper left corner of the image; in blue, *World Coordinates* (metres) centred in the middle of the detected palm. In orange, the *Real-World Coordinates* (metres) estimated by the GMH-D framework, centred in the RGB-D recording camera. For *Image Coordinate* of GMH, axis $Z_{im}$ expresses an adimensional depth parameter, relative to the wrist and scaled as the other two axes

Another study performed validation of GMH in the perspective of its usage for clinical evaluation of finger tapping and hand opening-closing movements [9]. The study also proposed an enhanced version of the framework exploiting an RGB-D camera (MAK), namely GMH-D. This framework has time performance comparable to GMH, but enhances the accuracy of the 3D tracking by leveraging both the depth estimation performed by the DL model and the depth-map provided by the RGB-D camera. Indeed, the depth estimation for each joint ($\hat{d}_j$, equation 1) is derived from the depth value of the wrist as measured by the on-board depth sensor ($d_{wrist}$) and the estimation done by the neural network ($z_{j,im}$, refer to Figure 1).

$$\hat{d}_j = d_{wrist} + z_{j,im} d_{wrist} \qquad (1)$$

The wrist was selected since it is the origin of the reference system employed by GMH and it represents the most stable joint, tracked within a surface much larger than that of the fingers. This permits to avoid errors in the retrieval of its depth value from the depth map provided by the sensor, which could depend on virtual marker misplacement by GMH or boundary interference in the depth sensors due to motion [9]. The Python code for running the GMH-D algorithm on videos captured by a MAK camera can be found at https://github.com/gianluca-amprimo/GMH-D.git. The authors validated the improvement provided by GMH-D over GMH by comparing the measurements of maximum and minimum peaks in the distance between relevant hand joints during finger tapping and hand opening-closing movements [9]. They achieved this by using a ruler positioned close to the hand of the subject performing the task, to retrieve real-world distances from video analysis. This preliminary validation, although achieving promising results, provides limited estimation of the quality of the tracking over the complete tasks, for which the support of a motion capture system is required.

Thus, the goal of this work is to further validate this previously obtained result, since GMH-D and GMH, due to their stability, easiness of deployment, and low-computational power seem to be a promising marker-less and non-invasive solution for clinical assessment of hand and fingers motion.

## 3. Setup and validation protocol

This section briefly explains the challenges of validating hand tracking from RGB and RGB-D videos simultaneously with a motion capture system. The proposed system setup and the protocol designed for validating GMH and GMH-D on three relevant clinical tasks to assess hand dexterity are then described.

### 3.1. Challenges of motion capture validation

Using motion capture for the validation of video-based HPE methods is a well-established practice [19, 36], since subject markerization does not excessively alter the appearance of the complete body shape nor impairs the motor performance of the subject. However, the same validation is more cumbersome in the case of hand tracking, especially from RGB-D cameras. Indeed, hands have a reduced surface, therefore the density of markers required to track all the degrees of freedom is much higher than that required for validating HPE methods.

Markerization inevitably alters the appearance of the hand, likely causing a reduction of accuracy in the DL model (Figure 2, left image). In the case of markered gloves, this impairment could even result in the complete impossibility to track the hand. Moreover, an excessive number of markers could reduce the dexterity of the hand, limiting the possible tasks to validate.

In addition, passive reflective markers for motion captures systems are complex to use in combination with RGB-D cameras, since they generate holes in the depth map provided by the sensor, as shown in Figure 2 (image on the right). Therefore, markers placement and recording camera viewing angle should be properly established such that virtual joints estimated by the DL model are not associated with such holes, leading to inaccurate estimation.

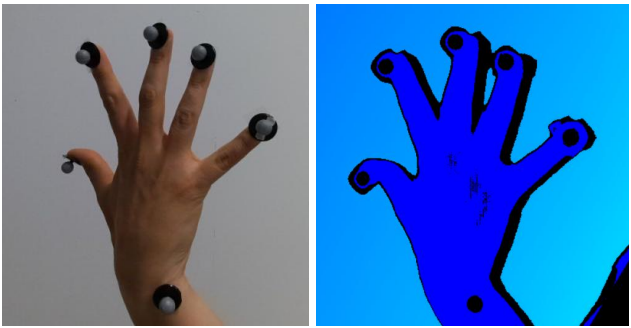

**Figure 2:** On the left, hand appearance when applying minimal markerization for motion capture (only the tips of fingers and the wrist reference); on the right, the same markered hand as seen by the depth sensor of the RGB-D camera. As it can be observed, passive-reflective markers produce holes in the estimated depthmap

Finally, time synchronisation between the motion capture system and the device running the DL framework should be managed to align the data collected, otherwise jitters between the recordings could generate results hard to objectively compare. In the validation of a system using RGB-D cameras, synchronisation is even more important, since both the motion capture system and such devices work in the infrared spectrum. The interference between infrared projectors may impair the quality of GMH-D, by producing frames with corrupted depth maps. With synchronisation, the sampling instants of the two systems are intertwined to suppress this unwanted effect.

## 3.2. Validation setup

Data acquisition sessions were organised at the Engineering for Health and Well-Being (EHW) Laboratory of the National Research Council (Institute of Electronics, Information Engineering and Telecommunications) in Turin, where a gold-standard motion capture system is available. The system consists of an Optitrack (OPT) solution including six Prime13 cameras having a resolution of 1280×1024 pixels. OPT cameras operate at 120 fps covering a working volume of approximately 6x4x3 $m^3$. The system was calibrated before each acquisition session, obtaining a residual value of 0.6 mm. The residual value is an average offset distance, in mm, between the converging rays when reconstructing a marker; hence, it is related to the reconstruction precision. The final estimated measurement error was less than 2.8 mm. Reflective markers of size 25mm were exploited. All tests were performed in the central zone of the recorded volume, thus ensuring maximum tracking accuracy. MAK was positioned in this area, stably fixed on a tripod (1 meter high), to capture videos of the participants' performance. MAK was connected to a laptop (Alienware m15 R2 I7-9750H, 16G RAM, NVIDIA GeForce RTX-2070 MaxQ with 8 GB di GDDR6). Figure 3 illustrates the complete experimental setup.

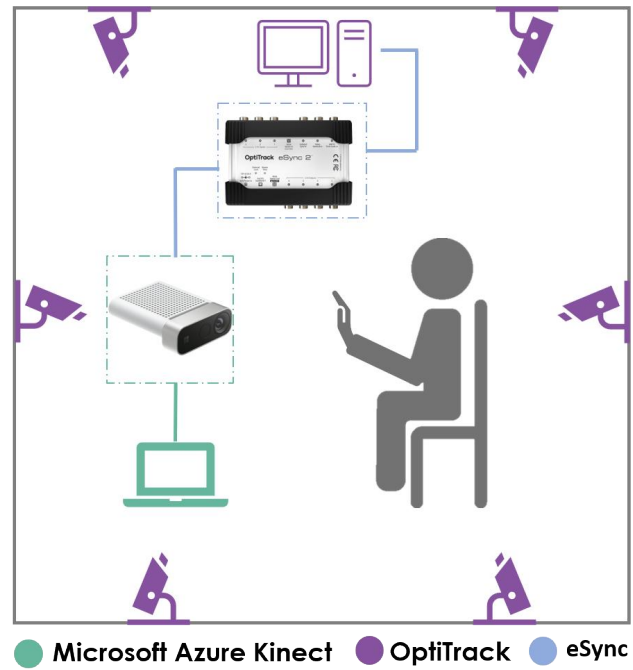

● Microsoft Azure Kinect ● OptiTrack ● eSync

**Figure 3:** Setup for the data acquisition sessions: an OptiTrack system (OPT) with 6 cameras is employed together with the Microsoft Azure Kinect DK (MAK). The Optitrack eSync 2 device is used for synchronisation. The subjects are told to seat in front of MAK in the centre of the working volume of the motion capture system and perform the assessment tasks.

The two systems were synchronised by mean of a sync generator (OptiTrack eSync2). The eSync2 was configured to operate as a *master* sync generator by driving MAK with a 30Hz sync signal and synchronising the OPT cameras at 120 fps using the internal 4x frequency multiplier.

## 3.3. Selected tasks and participants

To validate the performance of GMH and GMH-D, three exercises were considered: hand Opening and Closing (OC), Single Finger Tapping (SFT), and Multi Finger Tapping (MFT). OC consists of repeatedly opening and closing the hand. SFT consists of repeatedly tapping the thumb and index fingers. MFT consists of repeatedly and sequentially tapping the index, middle, ring, and little finger against the thumb.

These tasks are dynamically challenging and commonly used to measure fine hand dexterity and motor dysfunctions in the elderly and in pathological conditions. Therefore, they were considered to evaluate the tracking ability and accuracy of both frameworks [67, 89, 45, 20]. In particular, the SFT task is frequently addressed in works that apply DL hand tracking for Parkinson's disease diagnosis and staging through an automatic pipeline [48, 29, 10, 51, 102, 92, 98, 68]. In addition to the selected exercises, an initial Static Open Hand (SOH) phase was also recorded to extract participants' hand size.

As this study only aimed to validate the frameworks, healthy adult volunteers were involved. Specifically, ten subjects (4 females, 6 males), age $31.10 \pm 7.80$ years old were recruited. Average hand length (from middle finger tip to wrist, as retrieved from SOH task) for the male and female groups was $18.92 \pm 0.83$ cm and $16.5 \pm 0.78$ cm, respectively. None of the participants had physical hand/wrist/arm problems that could prevent them from performing the planned manual motor tasks. The experimental study was organised according to the Declaration of Helsinki (1964) and the lastest amendments, supervised by a clinician, and approved by the local Ethical Committee of Istituto Auxologico Italiano. Each participant signed an informed contents after receiving details on the study purposes and instrumentation.

### 3.4. Validation protocol

In the validation procedure, three significant influencing factors were investigated:

- **Distance from camera**: even though both methods perform a 3D estimation, the displacement from the camera position could affect how the underlying DL model identifies depth clues in the captured video.

- **Velocity of motion**: in low frame rate recording devices (<60 fps), motion blur alters the appearance of fingers [9]. This can produce inaccurate virtual marker positioning and distorted hand reconstruction.

- **Camera viewing angle**: especially in the case of SFT, different camera perspectives can modify the number of self-occluded joints during the execution of the movement.

The protocol was designed to verify performance and robustness of GMH and GMH-D with respect to the above mentioned factors and the type of performed task.

Regarding distance from camera, two ranges were considered: near distance, between 60 cm and 80 cm from MAK, and far distance, between 80 cm and 100 cm from MAK. This factor was studied in all tasks.

About velocity of motion, three different speed were investigated by coordinating the execution of the task with the rhythm of a metronome. For SFT and OC, low speed at 75 beats per minutes (bpm), normal speed at 115 bpm, and high speed at 140 bpm were selected. During the trials at different speeds, subjects were also asked to achieve different ranges of finger motion, compatible with the requested speed: slow speed-wide excursion, normal speed-free excursion, high speed-small excursion. This was done to ensure variability in the types of movements performed for the same task. For MFT, since the complexity of performing the movement correctly even for healthy subjects drastically increases with the speed, only a normal speed (115 bpm) was requested, and this confounding factor was therefore neglected.

Finally, the camera viewing angle was studied for SFT only, since both a lateral and a frontal perspective allow to observe the movement with different degrees of self-occlusion. In contrast, for OC and MFT, the only possible viewing angle is the frontal one, since in the lateral positioning of the camera the tracking of the palm by the PD model of GMH could be complex and likely to result in poor accuracy.

Table 1 summarises the protocol and the number of trials acquired for each task. Each recorded trials lasted 15 s. Eventually, a dataset composed of 200 videos was collected considering all the three tasks, plus 10 initial SOH trials.

Since hand markerization may significantly impact the quality of the tracking for both GMH and GMH-D, for each task a minimal markers configuration was selected, such that just the fundamental trajectories for assessing the task could be estimated and compared. Physical markers were placed on the back of the hand to avoid depth map holes and in close correspondence to the positions where virtual joints of GMH should lie, to limit the systematic positioning error between the two tracking systems and OPT. For SFT (Figure 4, left image), Wrist Outer Bone (WOB), Wrist Inner Bone (WIB), Index Finger Tip (IFT), and Thumb Tip (TT) were selected. Wrist markers provide a reference for the hand structure, while IFT and TT joints are those actively involved in the movement and their relative distance is usually employed in literature for the assessment of the task. For MFT (Figure 4, middle image), also Middle finger Tip (MT), Ring Finger Tip (RFT), and Pinkie Tip (PT) were marked, to evaluate the relative distance between all fingers tips and TT along the whole task. Finally, for OC and SOH (Figure 4, right image) just WOB, WIB, and MFT markers were applied.

## 4. Comparison methods and evaluation metrics

The comparison between GMH, GMH-D, and the OPT system focused on specific relative distances between finger joints deemed relevant for the objective assessment of the task. As already discussed, this allows to limit the number of required markers, thus avoiding to significantly alter the appearance of the tracked hand.

Moreover, focusing only on inter-finger distances also allows to avoid complex calibrations among the tracking systems. Indeed, whereas GMH-D is centred in the MAK colour camera (see Figure 1) and OPT can be calibrated to be aligned in this point as well, GMH *World coordinates* are centred in the palm of the tracked hand. Therefore, an additional calibration would be necessary to move this reference to the MAK centre. However, inter-finger distances are invariant to the translation and rotation between the reference systems, therefore they may be straightforwardly compared.

For OC, MT-Wrist Bone (WB) distance was evaluated, with WB as the middle point between WIB and WOB markers of OPT. Variations in this distance convey alterations in the closing of the hand. For SFT, the IFT-TT distance was evaluated. For MFT, IFT-TT, MT-TT, RFT-TT, and PT-TT distances were assessed, plus a virtual overall trajectory defined as the sum of these sub trajectories (TT-ALL). All the investigated distance trajectories are reported as dotted line in Figure 4.

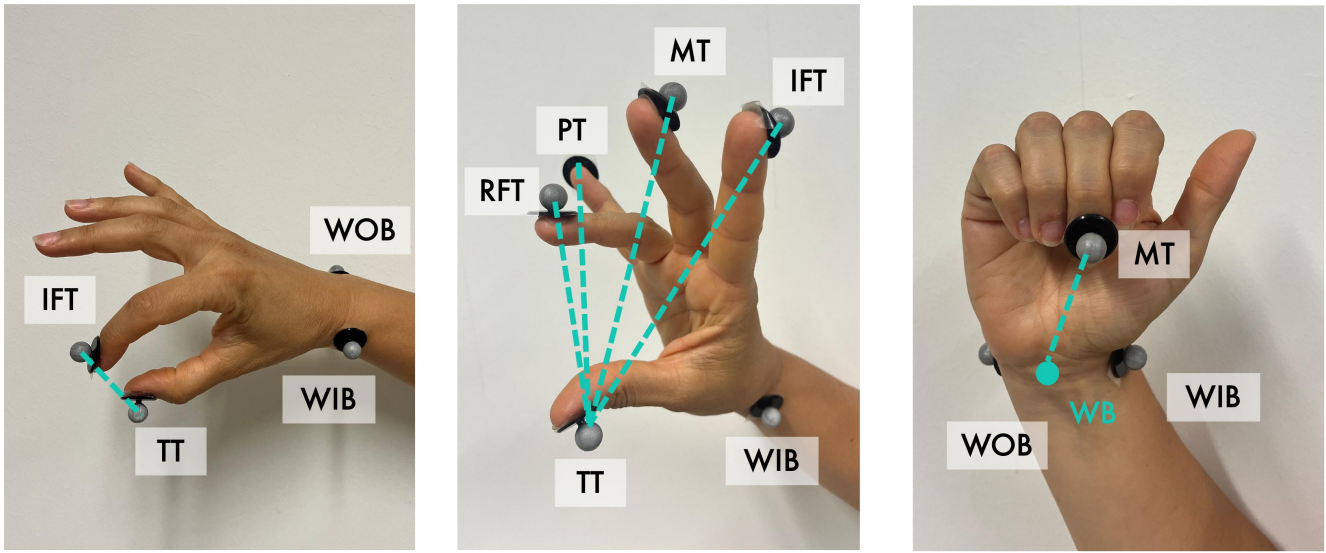

**Figure 4:** Markerization schema for the three tasks: on the left, markerization for SFT involves only wrist, index and thumb tips; in the centre, markerization for MFT involves all finger tips and the wrist; on the right, markerization for OC and SOH involves the wrist reference and the middle finger tip.

| | Distance | Viewing Angle | Speed | | | TOT |
|---|---|---|---|---|---|---|
| | | | Slow 75 bpm | Normal 115 bpm | Fast 140 bpm | |
| **OC** | Near 60-80 cm | Frontal | 20 | 10 | 20 | 50 |
| | Far 80-100 cm | Frontal | / | 10 | / | 10 |
| **SFT** | Near 60-80 cm | Lateral | 20 | 10 | 20 | 50 |
| | | Frontal | 20 | 20 | 20 | 60 |
| | Far 80-100 cm | Lateral | / | 10 | / | 10 |
| **MFT** | Near 60-80 cm | Frontal | / | 10 | / | 10 |
| | Far 80-100 cm | Frontal | / | 10 | / | 10 |
| **TOT** | | | 60 | 80 | 60 | 200 |

**Table 1**
To validate the two frameworks, a total of 200 videos lasting each around 15 s were recorded. Ten participants performed 3 different hand dexterity tasks: hand opening and closing (OC), single finger tapping (SFT) and multi-finger tapping (MFT). Requested speed (slow, normal, and fast) of execution, distance from the camera (near, far), and viewing angle (frontal, lateral) were modified in order to explore the performances of GMH and GMH-D as they vary. The number of videos recorded in each set-up is reported in the table as speed, distance, and viewing angle vary.

## 4.1. Whole-trajectory comparison

Before comparing the raw distances representing each task, the vertical offset due to misalignment between real and virtual markers and the horizontal offset due to residual time-shift between MAK and OPT were removed. To vertically realign the OPT trajectory with those of GMH-D and GMH, the mean distance between each point was evaluated and subtracted to OPT trajectory (physical markers lie on top of virtual markers being attached to the finger tips). The temporal realignment, instead, was performed using the cross-correlation method [2] and found to be almost constant in all videos due to optimal hardware synchronisation (6 frames delay between OPT and MAK recordings).

After this procedure, the selected distances as computed by OPT, GMH, and GMH-D were compared in terms of Root Mean Square Error (RMSE), which is a common error metric for comparing measurement systems. However, for a fairer comparison among trials with different excursions in the execution of the task, the Percentage Root Mean Square Error (PRMSE) was also computed. This metric is defined in Equation 2 as:

$$PRMSE = \sqrt{\frac{1}{n}\sum_{i=1}^{n}\left(\frac{Y_{OPT} - \hat{Y}}{Y_{OPT}}\right)^2} \times 100\% \qquad (2)$$

where $Y_{OPT}$ is the measurement obtained from the OPT system, whereas $\hat{Y}$ is the measurement estimated by either GMH or GMH-D.

In addition, the Pearson's correlation coefficient $\rho$ between the trajectory estimated by OPT and that estimated by either GMH or GMH-D was evaluated. This metric is related to the coherence in the trends measured by the different systems.

### 4.2. Single-segments comparison

To further verify the robustness of the two frameworks, a finer comparison was carried out by estimating the Range of Motion ($ROM$) in cm and the time duration ($DUR$) in seconds of single repetitions (movements) of the task, e.g. single taps of the TT and IFT joints in SFT. To do so, a segmentation algorithm was implemented in Matlab 2020b. Each movement segment was identified using local minima and maxima in the measured distances, which correspond to the minimum and maximum distance between reference markers of the motion. These two parameters are quite general; hence, they could be estimated for the segments of all the three tasks. In the case of MFT, the procedure was applied only to the TT-ALL distance, which summarises the motion of all involved fingers. The segmentation procedure was repeated identically for GMH, GMH-D, and OPT.

For each task, all the collected videos were considered, to achieve a dataset containing 1430, 1944, and 482 single segments of movement respectively for OC, SFT, and MFT. Segment-level parameters ($ROM$ and $DUR$) in each task were compared between OPT and GMH/GMH-D using Bland-Altman plot. Moreover, Lin's Concordance Correlation Coefficient (CCC) [55] and the Intraclass Correlation (ICC) [42] were estimated to measure the level of agreement between the two frameworks and OPT. In particular, CCC is an alternative version of ICC used to assess inter-rater variability, and is often employed to compare different measurement systems [43, 16]. It is defined in Equation 3 as

$$CCC = \frac{2\rho\sigma_x\sigma_y}{\sigma_x^2 + \sigma_y^2 + (\mu_x - \mu_y)^2} \qquad (3)$$

where $\rho$ is Pearson's correlation coefficient between random variables $X$ and $Y$ (either parameters from OPT and GMH or from OPT and GMH-D), $\mu$ and $\sigma$ are respectively the mean value and the standard deviation of the distributions of $X$ and $Y$. Following [8, 43, 16], a value in excess of 0.8 denotes high agreement between the two systems. For ICC, a threshold 0.75 is normally considered as a measure of high level agreement between different raters [42].

Finally, to provide a final overview of the tasks in the spectral domain, the dominant frequency of the voluntary movement spectral band ($F_{DOM}$) was identified together with its associated spectral power ($POW_{DOM}$). A comparison of $F_{DOM}$ and $POW_{DOM}$ as estimated by GMH, GMH-D and OPT in each trial was carried out using Bland-Altman plots, ICC, and CCC.

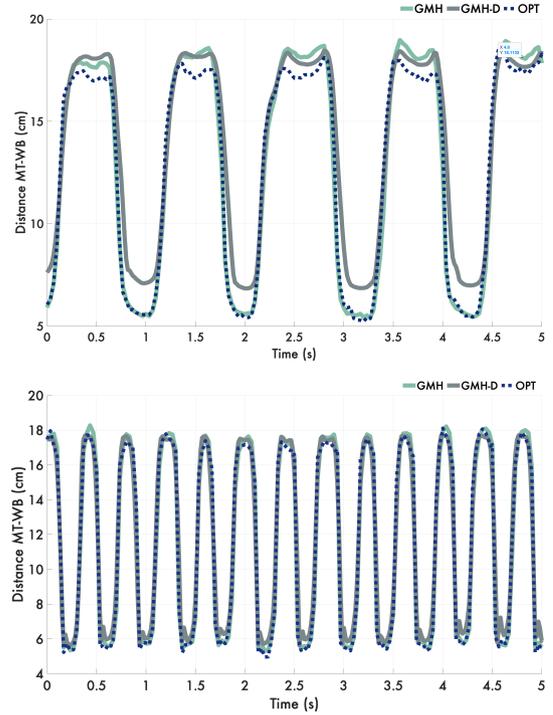

**Figure 5:** MT-WB distance during slow (top) and fast (bottom) OC trials as measured by GMH and GMH-D with respect to the gold standard OPT (dotted line). The three curves have been vertically and horizontally realigned to allow a direct comparison.

## 5. Results and discussion

The results of the validation procedure are organised according to the complexity of the motion to track. The OC task was considered the least complex to track because all fingers move together, followed by SFT, which involves the highly dynamic and coordinated movement of two fingers. Finally, MFT was deemed the most challenging as it involves the highly dynamic and coordinated movement of all fingers. A final discussion summarises the findings and provides guidelines on the application of the two frameworks for clinical assessment.

### 5.1. OC Validation

Figure 5 reports an example of the estimation of the MT-WB distance (OC task) performed by GMH and GMH-D with respect to the OPT gold standard measurement (dotted line), considering a slow (top) and a fast (bottom) execution. As it can be observed, both methods reconstruct with good precision the relative distance between the joints of interest, with perfect temporal alignment.

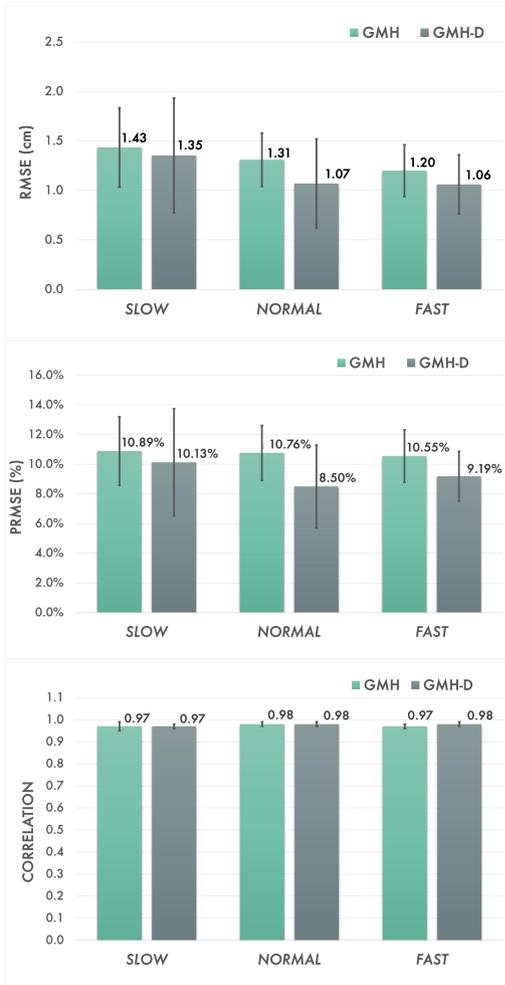

**Figure 6:** RMSE (top), PRMSE (middle), and Pearson's $\rho$ (bottom) bar plots with standard deviation, in trials at different velocity: Slow (75 bmp), Normal (115 bpm), Fast (140 bpm). Values reported close to the bars refer to the mean value.

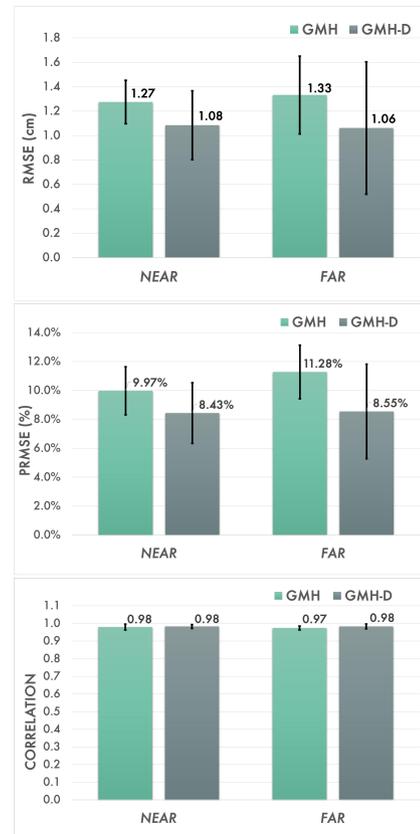

**Figure 7:** RMSE (top), PRMSE (middle) and Pearson's $\rho$ (bottom) bar plots with standard deviation, in OC trials at different distances from camera: near distance (60-80 cm) and far distance (80-100 cm). Values reported close to the bars refer to the mean value.

Results on RMSE, PRMSE, and Person's $\rho$ are organised separately according to velocity and distance factors, in Figure 6 and Figure 7 respectively. As it can be appreciated, both in terms of RMSE and PRMSE, GMH-D achieves a smaller error compared to GMH with respect to OPT, but the standard deviation turns out to be larger. However, the two methods are overall comparable, independently of distance and velocity. The mean error is around 10% in the PRMSE, without any straighforward correlation with the velocity of execution.

The correlation between 3D trajectories achieves very high values (above 0.97) with reduced standard deviation, independently of the distance from camera or the movement velocity. This underlines an excellent agreement between the trends measured by GMH and GMH-D and the gold standard (OPT). Regarding distance, from PRMSE we can observe that both standard deviations worsen with increasing distance, with also a slightly increase in its mean value for GMH, whereas GMH-D appears to be more stable.

Therefore, a slight influence between the distance from the camera and the tracking of the OC task seems to be present, especially for GMH. The correlation with respect to OPT, however, remains excellent for both frameworks, no matter of the distance considered.

From the segment-level analysis, the Bland-Altman plots for $ROM$ and $DUR$ are reported in Figure 8. As for frequency parameters referring to the whole task, only $POW_{DOM}$ is reported, since both methods achieve perfect concordance in the estimation of $F_{DOM}$, as supported also by the computation of ICC and CCC. As for $ROM$, it can be appreciated that 95.67% and 94.76% of estimations exhibit an error in the range [-2.55, 0.93] cm (mean: -0.81 cm) for GMH-D and [-2.87, 0.95] cm (mean: -0.96 cm) for GMH, respectively, with few outliers. The slightly narrower error range and the percentage of points included in it suggest slightly better performance in the estimation by GMH-D. The same holds true for $DUR$, where 96.02% of errors falls in the range [-0.12, 0.12] s (mean: 0.00 s) for GMH-D and 92.66% in the range [-0.14, 0.14] s (mean: 0.00 s) for GMH. As for $POW_{DOM}$, GMH exhibits a narrower error range than GMH-D, but this results is likely caused by some large outliers present in the data.

| | **GMH** | | | | | | **GMH-D** | | | | | |
|---|---|---|---|---|---|---|---|---|---|---|---|---|
| | | *ICC* | | | *CCC* | | | *ICC* | | | *CCC* | |
| | *Low Conf.* | **Value** | *High Conf.* | *Low Conf.* | **Value** | *High Conf.* | *Low Conf.* | **Value** | *High Conf.* | *Low Conf.* | **Value** | *High Conf.* |
| *ROM* | 0.79 | **0.81** | 0.82 | 0.68 | **0.71** | 0.72 | 0.88 | **0.89** | 0.90 | 0.80 | **0.82** | 0.83 |
| *DUR* | 0.96 | **0.96** | 0.97 | 0.96 | **0.96** | 0.97 | 0.97 | **0.97** | 0.98 | 0.97 | **0.97** | 0.98 |
| $F_{DOM}$ | 1.00 | **1.00** | 1.00 | 1.00 | **1.00** | 1.00 | 1.00 | **1.00** | 1.00 | 1.00 | **1.00** | 1.00 |
| $POW_{DOM}$ | 0.94 | **0.96** | 0.98 | 0.93 | **0.96** | 0.97 | 0.84 | **0.90** | 0.94 | 0.83 | **0.90** | 0.93 |

**Table 2**
ICC and CCC values for segment-level and frequency parameters in OC task, both for GMH and GMH-D methods with respect to the gold-standard OPT.

The results for ICC and CCC are reported in Table 2, using a 95% confidence level (low and high confidence ranges are reported). For ICC, *p-values* are all below $p < 0.001$ (not reported in the table for the sake of clarity). The ICC values highlight a high level of agreement (>0.96) for all the four investigated parameters, with almost perfect agreement for temporal and spectral properties, either using GMH or GMH-D. Slightly worse results, albeit still in excess of 0.8, are achieved for *ROM*, with GMH-D slightly outperforming GMH (0.89 vs 0.81). The CCC confirms these results, pointing out a larger discrepancy between *ROM* when exploiting GMH-D vs GMH, hence favouring the first method. This is above the threshold of 0.8, denoting an excellent agreement. It must be considered that the discrepancy of *ROM* is related to the 10% PRMSE average discrepancy between the trajectories, but could also be affected by the trivial segmentation algorithm. Indeed, how minimum and maximum points are identified depends on the morphology of the MT-WB distance measured by the GMH and GMH-D and, consequently, could have an impact on the evaluated parameters.

## 5.2. SFT validation

Figure 9 reports an example of the estimation of the IFT-TT distance by GMH and GMH-D with respect to the OPT gold standard (dotted line), considering a slow (top) and a fast (bottom) execution of the SFT task. In contrast with OC, it is evident that GMH suffers from a squeezing effects in the estimation of *ROM*, which is much more limited as an effect for GMH-D. This alteration is magnified by the increase in the execution speed, with an evident error in the spatial tracking of the IFT and TT joints when the two are in close contact. This effect was already identified in [9], where an error in the $z_{WO}$ component (refer to Figure 1) of IFT and TT was observed during the execution of the SFT task.

For SFT, the analysis of the velocity factor takes into consideration also the recording viewing angle (either lateral or frontal). The RMSE, the PRMSE, and the Person's $\rho$ values for the collected trials are shown in Figure 10. From the bar plots, the following considerations can be derived. A decrease in the error and a slightly better correlation is achieved when moving from the frontal to the lateral viewing angle for GMH-D, since this view possibly improves the evaluation of the depth by the MAK sensor. This is evident both from the mean values and the standard deviations. In addition, for GMH-D, the velocity of execution seems to have a marginal effect, especially considering PRMSE and the correlation. The slight change in RMSE is likely connected to the difference in the achieved *ROM* (wide vs small excursions) in the three type of trials, which does not affect instead PRMSE. Overall, the mean value for this metric falls in the range 8-15%.

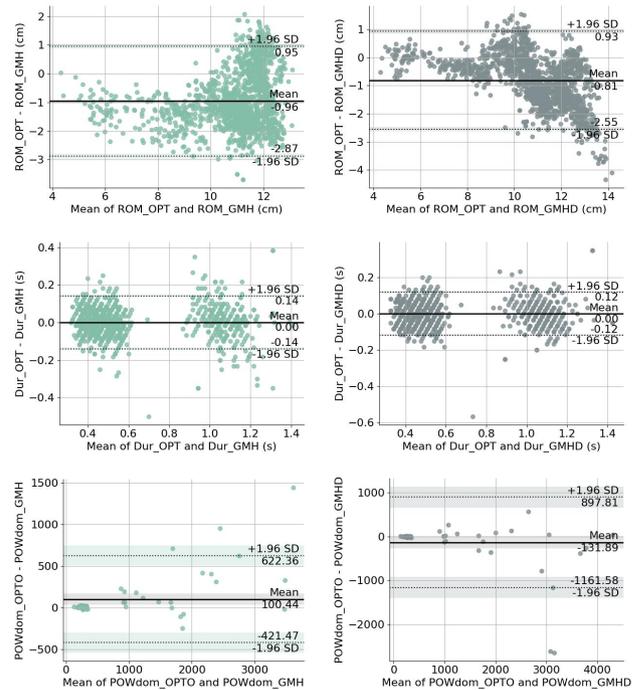

**Figure 8:** Bland-Altman plots for *ROM* (top), *DUR* (middle) and $POW_{DOM}$ (bottom) estimated from single repetitions of the OC task. Color coding for GMH and GMH-D is the same as in Figure 7.

Much greater error is measured for GMH both in the frontal and the lateral viewing angle (even four times more than that of GMH-D in the fast trials) and with a mean value much larger than 20% in all scenarios. Interestingly, PRMSE seems to amplify in the lateral viewing angle and steadily increases while increasing the speed of the SFT task. In the frontal viewing angle instead, the PRMSE seems less affected by the velocity factor, whereas correlation shows a reduction both in the frontal and lateral viewing angles. Overall, GMH-D seems to provide the best tracking quality in assessing the task, independently on the camera view.

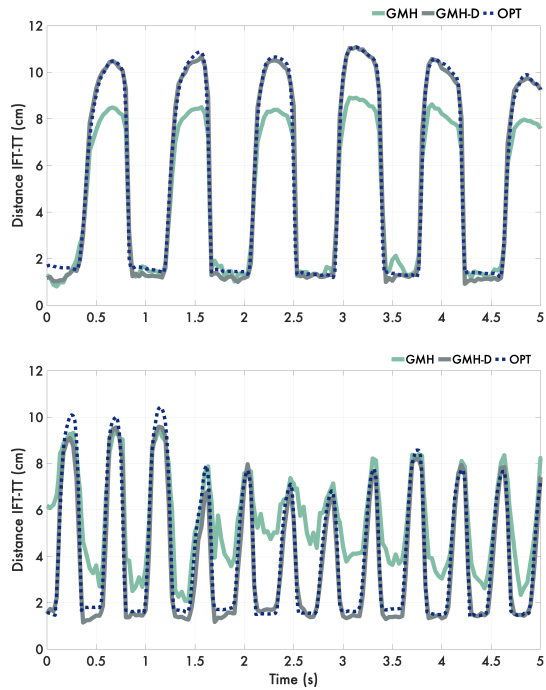

**Figure 9:** IFT-TT distance during slow (top) and fast (bottom) SFT trials, as measured by GMH and GMH-D with respect to the gold standard OPT (dotted line). The three curves have been vertically and horizontally realigned to allow a direct comparison.

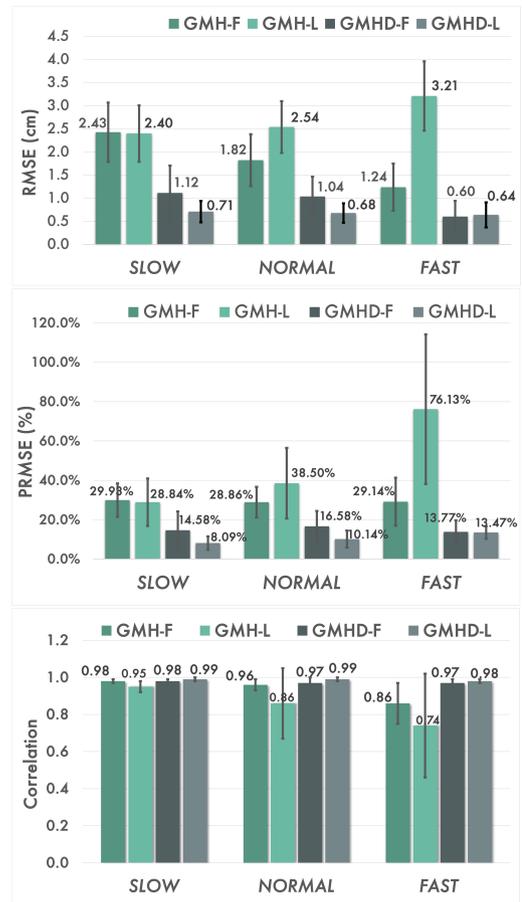

**Figure 10:** Bar plots of mean RMSE (top), PRMSE (middle), and Pearson's $\rho$ (bottom) with standard deviation, in trials at different velocity: low speed (75 bmp), normal speed (115 bpm), fast speed (140 bpm). Results are reported for the two studied recording viewing angle, either lateral (GMH-L, GMHD-L) or frontal (GMH-F, GMHD-F). Metrics mean values are written close to the bar to which they refer.

For the sake of brevity, the remaining part of the analysis focuses on the differences between GMH and GMH-D with respect to OPT only for the lateral viewing angle -i.e., the one for which the smallest errors (RMSE and PRMSE) and the largest correlations are achieved, considering all possible combinations of framework (GMH, GMH-D) and camera viewing angle (frontal or lateral).

First, the effect of distance from the camera is evaluated, by considering RMSE, PRMSE, and correlation. This result is reported in Figure 11.

The barplots further confirm the higher accuracy of GMH-D over the GMH framework, with a much smaller mean error and a reduced standard deviation. Moreover, GMH appears influenced by the distance from the recording camera, whereas GMH-D provides a tracking with a stable error at both the established distance ranges and with almost a perfect correlation -i.e, mean value above 98% and almost 0 standard deviation.

Moving to the segment-level analysis, the Bland-Altman plots for $ROM$, $DUR$, $F_{DOM}$, and $POW_{DOM}$ in Figure 12 are derived. As it can be observed, in 93.63% of the evaluations of $ROM$, the error between GMH-D and OPT lies in the range [-1.32, 1.51] cm (mean: 0.10 cm), whereas in 95.85% of estimation done by GMH have an error included in the range [0.69, 6.30] cm (mean: 3.49 cm) for GMH. This confirms the higher accuracy in estimating the 3D motion of the enhanced GMH-D framework, since most of the predictions either overestimate or underestimate the measure by around 1.5 cm, with a mean value of just 0.10 cm with respect to the 3.49 cm of GMH. For $DUR$, 95.07% of measurements have an error in the range [-0.15, 0.15] s (mean: 0.00 s) for GMH-D and 95.25% in the range [-0.19, 0.19] s (mean: 0.00 s) for GMH. Therefore performance are closer, but still GMH-D provides an higher accuracy. For the estimation of $F_{DOM}$, no error is performed by GMH-D with respect to OPT, whereas for GMH 91.53 % of estimations are producing an error between [-0.45, 0.62] Hz (mean: 0.08 Hz). Finally, for $POW_{DOM}$ 94.2% of estimations have an error in the range [-25.94, 18.85] (mean: -3.32) for GMH-D and 89.83% in the range [-86.21, 186.74] (mean: 50.27) for GMH, supporting again the higher accuracy of the first

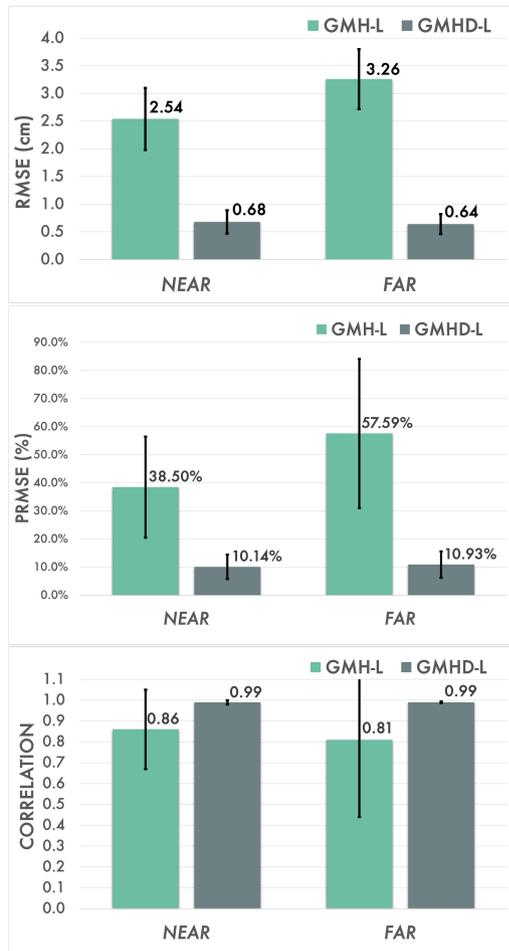

**Figure 11:** RMSE (top), PRMSE (middle) and Pearson's $\rho$ (bottom) bar plots with standard deviation, in SFT trials at different distance from recording camera: near distance (60-80 cm) and far distance (80-100 cm). Only lateral viewing angle is considered for GMH (GMH-L) and GMH-D (GMHD-L). Values reported close to the bars refer to the mean value.

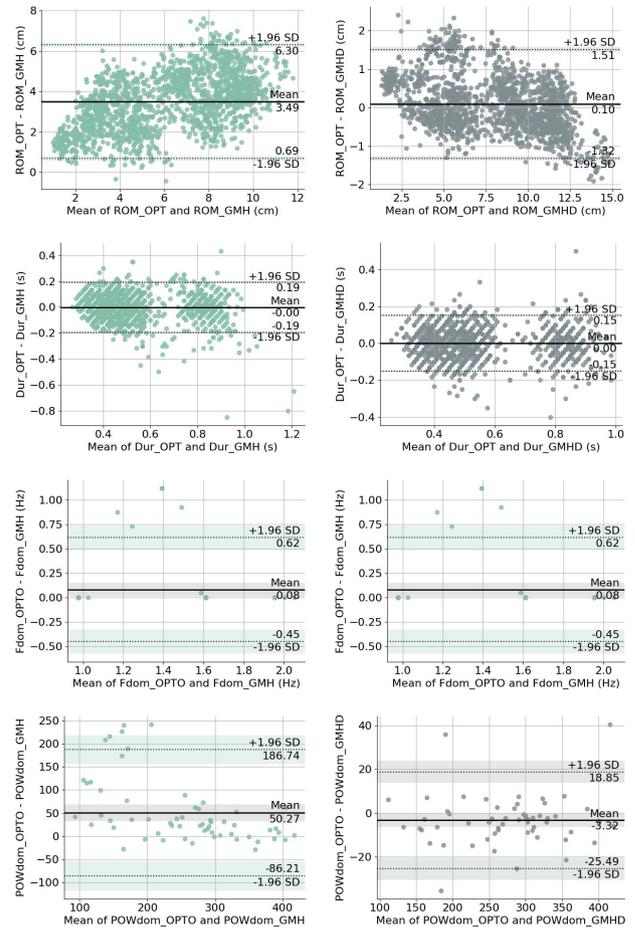

**Figure 12:** Bland-Altman plots for $ROM$ (top), $DUR$ (top-middle), $F_{DOM}$ (middle-bottom), $POW_{DOM}$ (bottom) estimated from single repetitions of the SFT task. Colour coding for GMH and GMH-D is the same as in Figure 7.

method also in terms of estimation of overall spectral properties of the movement.

The results for ICC and CCC are reported in Table 3, using a 95% confidence level (low confidence and high confidence levels are also reported). For ICC, *p-values* are all below $p < 0.001$, so they are not reported inside the table for the sake of clarity in table formatting. The values of ICC and CCC for GMH-D suggest an excellent level of agreement (>0.90) for all the four investigated parameters. In contrast, for GMH, it is confirmed how the method wrongly estimates $ROM$, producing measurements largely affected by errors. Nevertheless, a good level of agreement is achieved by the method to evaluate temporal and spectral parameters.

### 5.3. MFT validation

Figure 13 reports an example of the estimation of the RFT-TT distance (top) and the summarising distance TT-ALL (bottom) for MFT task, performed by GMH and GMH-D with respect to the OPT gold standard measurement

(dotted line). As it can be observed, the task is the most challenging for both DL tracking methods, that coherently follow the trend of the distance measured by OPT, but with smaller adherence to the true curve shape. Furthermore, it is worth noticing how the TT-ALL distance reflects, as hypothesised, the cumulative motion of all fingers, thus providing a way to observe from one single trajectory all the tapping movements from the fingers involved.

Results on RMSE, PRMSE, and Person's $\rho$ are organised according to the distance factor in Figure 14. The values are reported for the IFT-TT (INDEX), MT-TT (MIDDLE), RFT-TT (RING), PT-TT (PINKIE) distances that compose the final TT-ALL trajectory.

As it can be appreciated, when considering the near distance from the camera, GMH-D performs overall better than GMH, with a mean PRMSE value below 15% and a mean RMSE equal or smaller than 1 cm for all fingers. Values are similar among all fingers and seem to suggest a slightly lower mean error for pinkie, probably due to the fact that this finger is always clearly visible in the task, making its tracking less ambiguous. Both GMH and GMH-D achieve a correlation value with OPT above 0.8. In the far condition, both GMH and GMH-D show an increase in the mean error and its standard deviation, and a small decrease in the correlation, suggesting an influence of the distance factor in the quality of the final reconstruction of the MFT task.

| | GMH | | | | | | GMH-D | | | | | |
| | ICC | | | CCC | | | ICC | | | CCC | | |
| | *Low Conf.* | **Value** | *High Conf.* | *Low Conf.* | **Value** | *High Conf.* | *Low Conf.* | **Value** | *High Conf.* | *Low Conf.* | **Value** | *High Conf.* |
| *ROM* | 0.86 | **0.87** | 0.88 | 0.47 | **0.49** | 0.51 | 0.97 | **0.98** | 0.98 | 0.97 | **0.98** | 0.98 |
| *DUR* | 0.84 | **0.85** | 0.86 | 0.84 | **0.85** | 0.86 | 0.89 | **0.90** | 0.91 | 0.89 | **0.90** | 0.91 |
| $F_{DOM}$ | 0.68 | **0.80** | 0.87 | 0.66 | **0.86** | 0.87 | 1.00 | **1.00** | 1.00 | 1.00 | **1.00** | 1.00 |
| $POW_{DOM}$ | 0.59 | **0.74** | 0.84 | 0.51 | **0.65** | 0.75 | 0.98 | **0.99** | 0.99 | 0.98 | **0.99** | 0.99 |

**Table 3**
ICC and CCC values for segment-level and frequency parameters in SFT task, both for GMH and GMH-D methods with respect to the gold-standard OPT, considering a 95% confidence interval

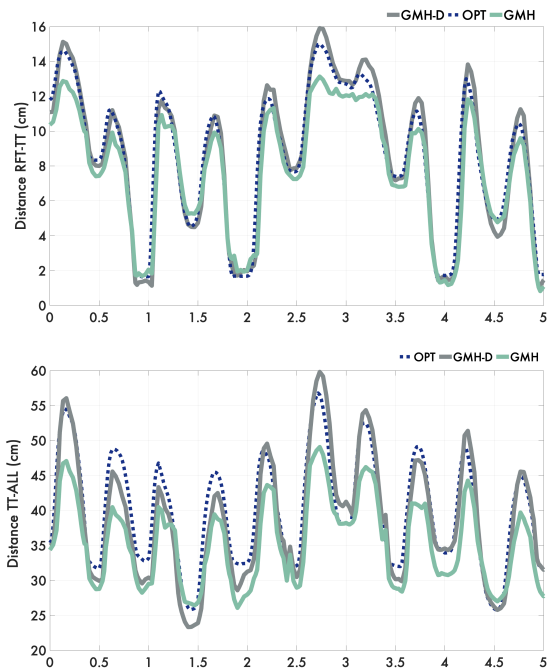

**Figure 13:** RFT-TT distance (top) and TT-ALL (bottom) distance during a MFT task, as measured by GMH and GMH-D with respect to the gold standard OPT (dotted line). The three curves have been vertically and horizontally realigned to allow a direct comparison.

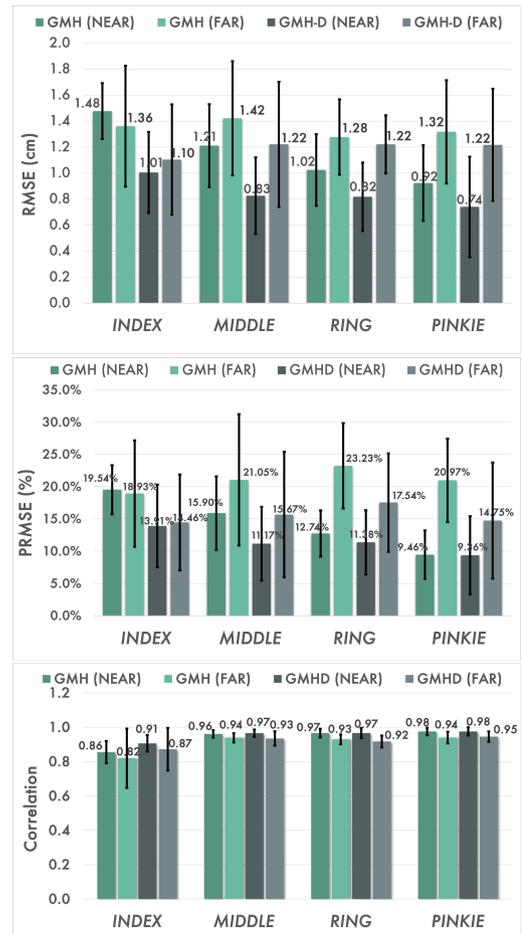

**Figure 14:** RMSE (top), PRMSE (middle), and Pearson's $\rho$ (bottom) bar plots with standard deviation, in MFT trials at different distances from camera: near distance (60-80 cm) and far distance (80-100 cm). Values reported close to the bars refer to the mean value.

Moving to the segment-level analysis from the TT-ALL distance, the Bland-Altman plots for *ROM* and *DUR*, $F_{DOM}$ and $POW_{DOM}$ in Figure 15 are shown. As a first remark, it must be noted the the mean value error and the value of the limits are inevitably larger due to the propagation in TT-ALL, either estimated by GMH or GMH-D, of the error present in the tracking of each inter-finger distance that is composing this fictitious trajectory. Therefore, an error range four times larger than that achieved in SFT was expected and observed. Proceeding with the analysis, as it

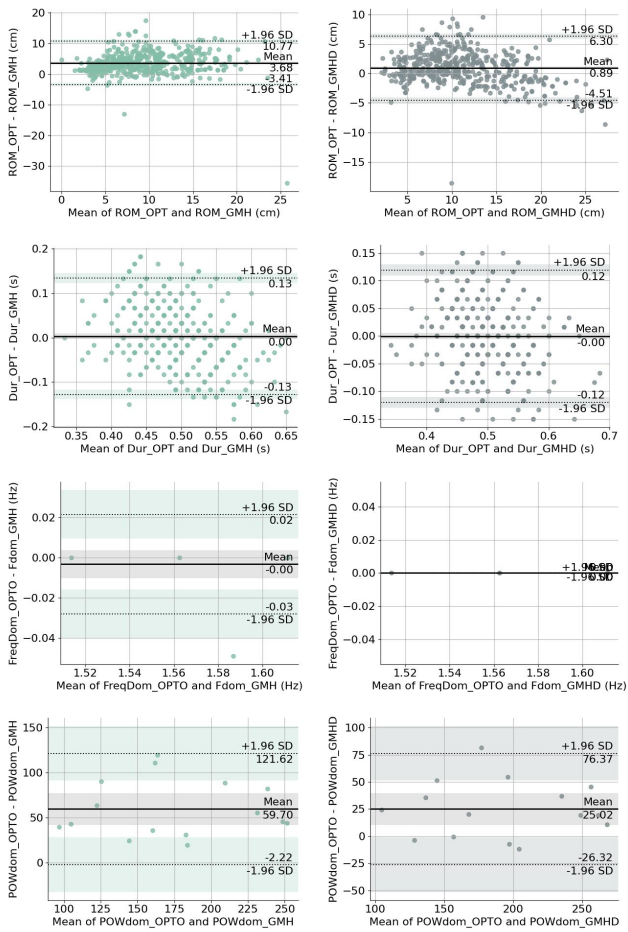

**Figure 15:** Bland-Altman plots for *ROM* (top), *DUR* (top-middle), $F_{DOM}$ (middle-bottom), $POW_{DOM}$ (bottom) estimated from single segments of the TT-ALL distance in MFT task. Colour coding for GMH and GMH-D is the same as in Figure 7

The results for ICC and CCC are reported in Table 4, using a 95% confidence level (low confidence and high confidence ranges are also reported). For ICC, *p-values* are all below $p < 0.001$ so are not reported for the sake of clarity in the table presentation. First of all, it must be noted that, due to the inclusion of trials with just one possible execution speed, the internal variability in terms of duration is really reduced. This, combined with the limited number of samples coming from one single type of trials (only one speed was tested with respect to the other tasks), produces low values of ICC and CCC for *DUR* parameters, either for GMH and GMH-D. However, from Bland-Altman plots, we can observe that actually almost all points are falling in a narrow error range, comparable with that achieved in the previous tasks. Therefore, the validity of these two metrics for *DUR* parameter is really limited since the inter-variability (between GMH/GMH-D and OPT) and the intra-variability of the single segments in terms of duration in the dataset are unbalanced- i.e., even mistakes with small magnitude produce a variability larger than the internal variability of *DUR*, biasing the results. Moreover, it must also be considered that, as already mentioned, the TT-ALL distance, due to its definition, includes and magnifies all the errors verifying in the tracking of trajectory of all the fingers involved, which can alter significantly the morphology of the curve (e.g., Figure 13) and the subsequent segmentation procedure, with evident effects on the estimation of the *DUR* parameter for the single segments. The values of ICC suggest a good level of agreement (>0.80) for all the remaining investigated parameters, with almost a perfect agreement for temporal and spectral properties either using GMH or GMH-D. Again, GMH-D outperforms GMH in terms of accuracy of the estimation of *ROM* (for ICC: 0.86 vs 0.71; for CCC: 0.85 vs 0.70).

### 5.4. GMH vs GMH-D: remarks

From the results achieved in the validation of the three proposed tasks, it is evident how both the RGB framework (GMH) and the RGB-D enhanced framework (GMH-D) hold potentiality in their application to track clinical assessment tasks. Overall, GMH-D seems to provide an accurate reconstruction of the motion in all the scenarios, with a mean PRMSE always smaller than 15% and an almost perfect correlation (above 0.9) between the tracked inter-finger distances in all the tasks, even the complex multi-finger tapping test. Moreover, this framework appears robust to different speed during the tasks and also to the distance from the recording camera, at least in the range 60 cm - 100 cm. The latter clearly depends also on the quality of the depth sensor of the RGB-D camera and was achieved using a MAK, which has an high accuracy for depth tracking up to 3.5 m [91]. When adapting GMH-D to other RGB-D devices, the quality of their depth sensors may clearly affect the result and should be evaluated beforehand. Still, this method can struggle with self-occlusions of fingers, as observed in SFT, where the lateral viewing angle improves the quality of tracking over the frontal one. Therefore, when can be observed, in 94.62% of the evaluations of *ROM*, the error between GMH-D and *OPT* lies in the range [-4.50, 6.50] cm (mean: 0.89 cm), whereas in 96.48% of estimations done by GMH have an error included in the range [-3.40, 10.76] cm (mean: 3.68 cm), confirming the higher accuracy in estimating the 3D motion of GMH-D over GMH. For *DUR*, 95.45% of measurements have an error in the range [-0.12, 0.12] s (mean: 0.00 s) for GMH-D and 95.65% in the range [-0.13, 0.13] s (mean: 0.00 s) for GMH, therefore performance are closer, but still GMH-D provides an higher accuracy both in spatial and temporal domains. Regarding the estimation of $F_{DOM}$, no error is performed by GMH-D with respect to OPT, whereas for GMH 88.89 % of estimations are producing an error between [-0.03, 0.02] Hz (mean: 0.00 Hz), almost negligible. Finally, for $POW_{DOM}$ 93.75% of estimations have an error in the range [-2.22, 121.62] (mean: 58.75) for GMH-D and 93.75% in the range [-26.32, 76.37] (mean: 25.02) for GMH, supporting again the higher accuracy of the first method also in terms of estimation of overall spectral properties of the movement.

| | GMH | | | | | | GMH-D | | | | | |
| | ICC | | | CCC | | | ICC | | | CCC | | |
| | Low Conf. | Value | High Conf. | Low Conf. | Value | High Conf. | Low Conf. | Value | High Conf. | Low Conf. | Value | High Conf. |
|---|---|---|---|---|---|---|---|---|---|---|---|---|
| *ROM* | 0.67 | **0.71** | 0.76 | 0.52 | **0.70** | 0.76 | 0.84 | **0.86** | 0.89 | 0.83 | **0.85** | 0.86 |
| *DUR* | 0.36 | **0.44** | 0.51 | 0.38 | **0.44** | 0.46 | 0.39 | **0.46** | 0.53 | 0.42 | **0.46** | 0.48 |
| $F_{DOM}$ | 0.88 | **0.90** | 0.93 | 0.84 | **0.89** | 0.93 | 1.00 | **1.00** | 1.00 | 1.00 | **1.00** | 1.00 |
| $POW_{DOM}$ | 0.57 | **0.83** | 0.85 | 0.73 | **0.75** | 0.78 | 0.69 | **0.88** | 0.96 | 0.81 | **0.82** | 0.84 |

**Table 4**
ICC and CCC values for segment-level and frequency parameters in MFT task, both for GMH and GMH-D methods with respect to the gold-standard OPT applying this kind of approach also the proper and optimal positioning of the hand should be taken into consideration. Overall, GMH-D allows to evaluate, with a good level of agreement, the four investigated parameter, namely *ROM*, *DUR*, $F_{DOM}$, and $POW_{DOM}$, thus allowing not only to reconstruct the movement, but also to derive from it some relevant parameters for clinical evaluation.

On the other hand, GMH appeared promising in the tracking of the OC task, with accuracy in the tracking of the MT-WB distance comparable to GMH-D. However, the analysis of the SFT and MFT highlighted how this framework fails in precisely reconstructing motion in 3D when finer movements are studied, producing very large RMSE and PRMSE values. From the analysis of trials with high error, it is evident how the model correctly marks the hand in 2D, but due to self-occlusions of joints, fails in properly reconstructing the 3D shape of the hand. For instance, in the SFT task, with lateral viewing angle recording, when the fingers are touching, GMH often does not identify the two finger tips as aligned over the z-axis, but misplaces one in front of the other, producing a wrong value in the minimum of the IFT-TT distance.

Overall, trajectory reconstructed by SFT and MFT seems to be affected by a squeezing effect due to the wrong estimation of depth by the model working on RGB video only. This behaviour is less observed in OC. This could be due to the fact that the OC task involves all fingers. Therefore, it may be more trivial for the DL model to reconstruct the motion of the whole hand. In addition, such model might lack training in the reconstruction of specific and finer hand configurations such as SFT and MFT, with respect to the more common opening-closing gesture. This failure in estimating properly depth would explain also the dependence of GMH from the distance from camera, with a worsening of the performances in the far positioning, and with increasing velocity of motion, which alters significantly the shape of the hand due to the motion blur. These issues clearly are cancelled or largely reduced by GMH-D thanks to the usage of the depth sensor. Moreover, it must be considered that GMH, as clearly stated by developers in [3], was not designed for this specific type of application and was mainly thought for working with close hand recordings, such as egocentric video from smartphone videocamera.

Nevertheless, even if GMH seems to be trustworthy in spatial analysis of motion only for OC task, with a good level of agreement with OPT in estimating *ROM*, it could still be applied to estimate temporal and spectral properties also in the other two tasks. Indeed, the framework shows good to excellent agreement in terms of ICC and CCC for *DUR*, $F_{DOM}$, and $POW_{DOM}$.

Summing up, GMH-D requires a slightly more complex hardware than GMH (RGB-D vs RGB camera) but can significantly improve the quality of the measurement, providing trustworthy information about ranges of motion in fine-grained 3D movements, with a good to excellent level of agreement with respect to the gold standard.

On the other hand, GMH has limited capability in 3D reconstruction, but thanks to the high quality of the 2D tracking of hand joints can still provide meaningful information about temporal and spectral property of motion. This finding is coherent with the results of the validation carried out for Parkinson's disease tremor evaluation in [33].

It must also be considered that some factors can have slightly enlarged the measured error values obtained for both frameworks: the residual offset between physical and virtual markers that likely was not completely removed by the pre-processing; the alteration of the appearance of the hand due to the physical markers, that caused in some cases a wrong positioning of the GMH virtual marker; for segment-level analysis, the trivial segmentation algorithm, which inevitably can introduce an additional error in the evaluation of ICC and CCC between parameters.

Unfortunately, due to the lack of comparable validation procedures for 3D RGB or RGB-D hand tracking methods based on DL in the literature, it is not possible to carry out a proper comparison with other studies. However, the results obtained are promising and demonstrate the potential, strengths, and weaknesses of the two DL-based frameworks, especially from the perspective of use in clinical applications.

# 6. Conclusions

This paper presented the validation against a gold standard system for motion capture of two DL-based hand tracking frameworks, namely Google MediaPipe Hand (GMH) and its enhanced version GMH-D. This validation was especially focused on the usage of such frameworks in clinical applications, such as automatic motion quality assessment.

This work aimed also at remarking the importance of carrying out a rigorous validation of DL-based methods before their application in scenarios such as clinical practice. Indeed, this allows to investigate exactly if a general DL solution could be meaningful and trustworthy for a clinical use case and whether it could properly measure parameters of interest. This is especially relevant taking into account that most *off-the-shelf* DL solutions for hand tracking are not specifically designed for deployment in clinical applications and could not generalise well to this scenario.

The validation of GMH and GMH-D was achieved by studying the tracking quality in three tasks commonly administered in clinical practice, namely Hand Opening and Closing, Single Finger Tapping, and Multiple Finger Tapping, characterised by different level of motion complexity. Three possible influencing factors (distance from recording camera, recording camera viewing angle, and velocity of tracked motion) were investigated. Concordance between the gold standard and the DL methods was evaluated using commonly estimated error metrics (RMSE, PRMSE, and Pearson's correlation) and by deriving, from tracked interfinger distances, some relevant parameters (range of motion, duration of single motion repetitions, dominant frequency of motion and its power). The agreement between such parameters was investigated using well-established metrics such as ICC and CCC.

Results suggest that for a more accurate reconstruction of 3D motion, RGB-D input is required and can provide results with good to excellent level of agreement to gold standard. RGB only input, even if less accurate and affected by specific factors, such as distance from camera and velocity of motion, can still be employed for evaluating temporal and spectral properties of motion with a good level of trust.

As a limitation, the validation did not take into account different light conditions that could alter tracking for both the RGB and RGB-D frameworks. However, since also the gold standard of motion capture has strict requirements on light conditions to function properly, how to carry out this evaluation is still an open challenge to investigate as future direction. Moreover, motions involving rotation of the hand, such as the prono-supination task, were not yet investigated, but could represent an interesting and additional challenging scenario for both methods. This task will be the focus of future investigation, to provide a more comprehensive evaluation of hand functionalities.

To conclude, thanks to the rapid growth of solutions for accurate hand tracking from RGB and RGB-D videos, a lot of new possibilities will arise in the next future for hand tracking and its usage in clinical applications. In this scenario, providing rigorous validation will become crucial to prove the efficacy and the trustworthiness of such methods, which is the direction to which this work is providing its main contribution.

*Code availability statement*  The code to run both GMH and GMH-D on videos acquired using Microsoft Azure Kinect is available in GitHub:

https://github.com/gianluca-amprimo/GMH-D.git.

*No competing interests statement*  The authors declare that they have no known competing financial interests or personal relationships that could have appeared to influence the work reported in this paper.

# CRediT authorship contribution statement

**Gianluca Amprimo:** Conceptualization, Methodology, Software, Validation, Formal Analysis, Investigation, Writing - Original Draft, Writing - Review & Editing, Visualization. **Giulia Masi:** Software, Formal Analysis, Investigation, Writing - Review & Editing. **Giuseppe Pettiti:** Methodology, Resources, Supervision, Writing - Review & Editing, Funding acquisition. **Gabriella Olmo:** Conceptualization, Writing - Review & Editing, Supervision. **Lorenzo Priano:** Writing - Review & Editing, Supervision. **Claudia Ferraris:** Conceptualization, Methodology, Resources, Validation, Formal Analysis, Investigation, Writing - Original Draft, Writing - Review & Editing, Supervision, Funding acquisition.